\documentclass[sigconf]{acmart}
\renewcommand\footnotetextcopyrightpermission[1]{} % removes footnote with conference information in first column
\usepackage{booktabs} % For formal tables
\usepackage{comment}
\usepackage[disable]{todonotes}   % notes showed
\usepackage{tikz}
\setcopyright{rightsretained}
\acmDOI{}

\acmISBN{}

\acmConference[Preprint]{Preprint}{March}{2018}
\acmYear{2018}
\copyrightyear{2018}

\begin{document}
\title{Generative Design in Minecraft (GDMC)}

\subtitle{Settlement Generation Competition}

\author{Christoph Salge}
\orcid{1234-5678-9012}
\affiliation{%
  \institution{School of Computer Science, University of Hertfordshire}
  \streetaddress{College Lane}
  \city{Hatfield}
  \state{UK}
  \postcode{AL10 9AB}
}
\affiliation{%
  \institution{Department of Computer Science and
Engineering,\\ New York University}
  \streetaddress{5 Metrotech Center}
  \city{Brooklyn}
  \state{NY}
  \postcode{11201}
}
\email{ChristophSalge@gmail.com}

\author{Michael Cerny Green}
\affiliation{%
  \institution{Department of Computer Science and
Engineering,\\ New York University}
  \streetaddress{5 Metrotech Center}
  \city{Brooklyn}
  \state{NY}
  \postcode{11201}
}
\email{mcgreentn@gmail.com}

\author{Rodgrigo Canaan}
\affiliation{%
  \institution{Department of Computer Science and
Engineering,\\ New York University}
  \streetaddress{5 Metrotech Center}
  \city{Brooklyn}
  \state{NY}
  \postcode{11201}
}
\email{rodrigo.canaan@nyu.edu}

\author{Julian Togelius}
\affiliation{%
  \institution{Department of Computer Science and
Engineering,\\ New York University}
  \streetaddress{5 Metrotech Center}
  \city{Brooklyn}
  \state{NY}
  \postcode{11201}
}
\email{julian@togelius.com}

%The default list of authors is too long for headers.
\renewcommand{\shortauthors}{C. Salge et al.}

\begin{abstract}

This paper introduces the settlement generation competition for Minecraft, the first part of the Generative Design in Minecraft challenge. The settlement generation competition is about creating Artificial Intelligence (AI) agents that can produce functional, aesthetically appealing and believable settlements adapted to a given Minecraft map --- ideally at a level that can compete with human created designs. The aim of the competition is to advance procedural content generation for games, especially in overcoming the challenges of adaptive and holistic PCG. The paper introduces the technical details of the challenge, but mostly focuses on what challenges this competition provides and why they are scientifically relevant.

\end{abstract}





\maketitle

\section{Introduction}
%k\subsection{Settlement Generation Challenge}

This paper introduces the settlement generation challenge\footnote{More info at: http://gendesignmc.engineering.nyu.edu/}. The task is to write an algorithm\footnote{also interchangeably referred to as an \textit{agent} or \textit{artificial intelligence}(AI)} that can take a given Minecraft~\citep{game:Minecraft} map and, without supervision, generate a settlement on this map by placing and deleting appropriate blocks. After the algorithm is submitted to the competition, it will be run on several previously unseen maps. The resulting settlements on the maps will then be evaluated by a panel of judges based on criteria outlined later in this paper. The team that builds the algorithm with the best average evaluation score wins.  

\begin{figure*}[tbh]
\includegraphics[width=\linewidth]{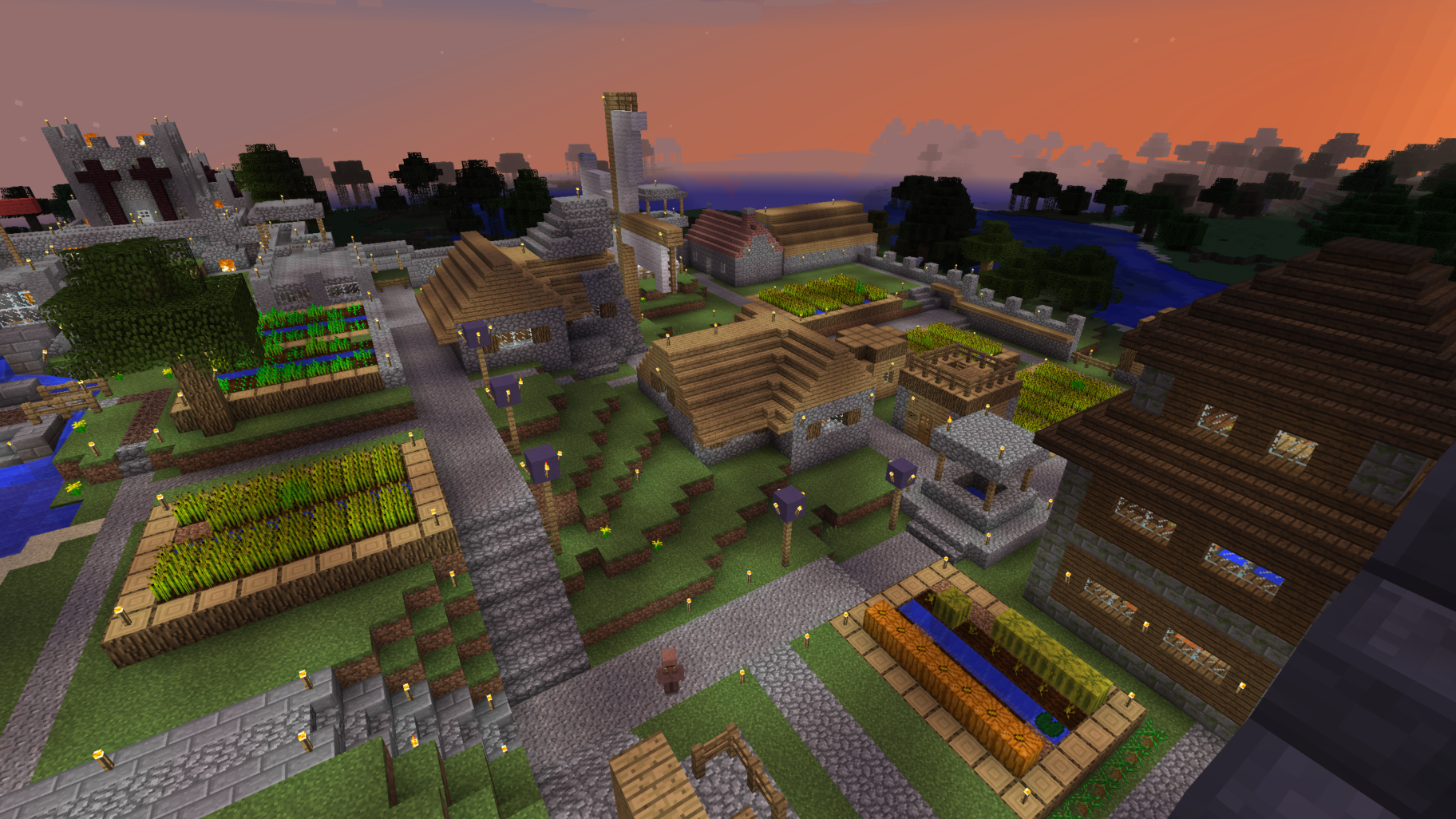}
  \caption{An example of a settlement in Minecraft that features both automatically generated and player generated elements. Most of the fields, and several of the houses in the center of the picture were generated by the game's settlement generation algorithm during map generation. All additional elements and modifications were added by players. }\label{fig:exampleSettlement}
\end{figure*}

\subsection{Overview}

The overall paper will give more details on the exact task, the framework to build the algorithm in and the evaluation criteria. The focus of this paper is to explain why we set up this challenge, and what we hope to achieve with it. To this end, we will first talk about motivations behind GDMC and introduce the concepts of adaptive and holistic PCG, which we believe this competition will advance. We will then discuss the particular challenges in settlement generation in Minecraft, and how they inform our evaluation criteria. We will also introduce the framework the agent should be written in and discuss the competition setup. The background section then describes some existing approaches that can serve as a starting point to build a settlement generator. Finally, we give an outlook on where we think this competition will be going in the future. 

\subsection{Motivation}

Computer science has a tradition of designing competitions which focus development towards overcoming specific goals. Examples include the DARPA Challenges, both for self driving cars \cite{buehler2009darpa} and bipedal robots \cite{atkeson2016happened}. Another example is Robocup \cite{kitano1997robocup}, a competition about robots playing soccer which led to developments in bipedal walking and team coordination. 

Artificial Intelligence (AI), in particular, has a rich history of prizes and competitions. Games are a popular topic here, as they usually have well-defined goals, scoring and winning conditions, which enable the use of many popular techniques such as tree search, reinforcement learning, evolutionary optimization and neural networks~\cite{togelius2016run}. 

However, not all games have clear goals, and not all AI is about winning a game. Take for example the popular game Minecraft \citep{game:Minecraft}, where players find themselves in a world made of three-dimensional cubes of earth, wood, lava and other materials. First steps usually include punching at a tree to obtain some wood, which is then used to build tools to mine other blocks. While it is technically possible to beat the game by defeating an end boss, most players entertain themselves with building structures and obtaining stuff from the world. There are enemies in the game, and other environmental hazards that can kill the player's avatar, but it is possible to largely avoid violence and focus on building interesting structures. The game has been compared to a digital version of Lego \cite{jimrossignol2010}, as it allows the players to use blocks to build a nearly infinite variety of structures. Building an artificial intelligence (AI) agent to play Minecraft \textit{per se} is therefore a rather ill defined task, as it is hard to define, even to humans, what playing Minecraft \textit{well} means. Nevertheless, Minecraft offers a complex interactive environment that can be used for AI testing \cite{johnson2016malmo,aluru2015minecraft}, given that an appropriate task has been defined \cite{katjahofmann2017}. 

One common human activity in Minecraft is to create structures that resemble settlements. Houses and lighting offer actual game play advantages, but villages and buildings are also popular forms of creative expression. Understanding this process of creative expression from a computational standpoint, i.e. building an AI that can mimic it, is a challenging undertaking. Nevertheless, we believe that understanding creativity and how to generate creative artifacts is an important part of intelligence \cite{colton2012computational}. We choose the task of settlement generation because it offers a relatively well contained problem, with somewhat understood evaluation criteria, that nevertheless offers a large space of possible solutions. 

We hope that our challenge will garner attention to  AI problems in areas referred to as procedural content generation \cite{shaker2016procedural} (PCG) or generative design \cite{mccormack2004generative}, which are both focused on using algorithms to design artifacts or create content. One problem with setting up a competition involving any kind of computational creativity is the lack of formalized evaluation criteria. Other existing PCG competitions, such as the GVGAI level generation competition \cite{khalifa2016general}, the GVGAI rule generation competition \cite{khalifa2017general} and the Mario AI level generation competition \cite{shaker20112010} all rely on human judges for evaluation. Consequently, we will also employ human judges, drawn from or nominated by a panel of experts.

In contrast to existing competitions, which employ a one dimensional preference ranking, i.e. ``Which of these two levels do you prefer?'', we plan to evaluate generated settlements in several categories, based on a set of criteria. We choose to employ a rather large set of criteria to ensure that the different facets of the problem are considered, and to force participants to find an approach balances a range of different requirements. Furthermore, we want to encourage solutions with a greater amount of adaptivity, so we designed criteria to encourage this.  

\begin{figure}[tbh]
\includegraphics[width=\linewidth]{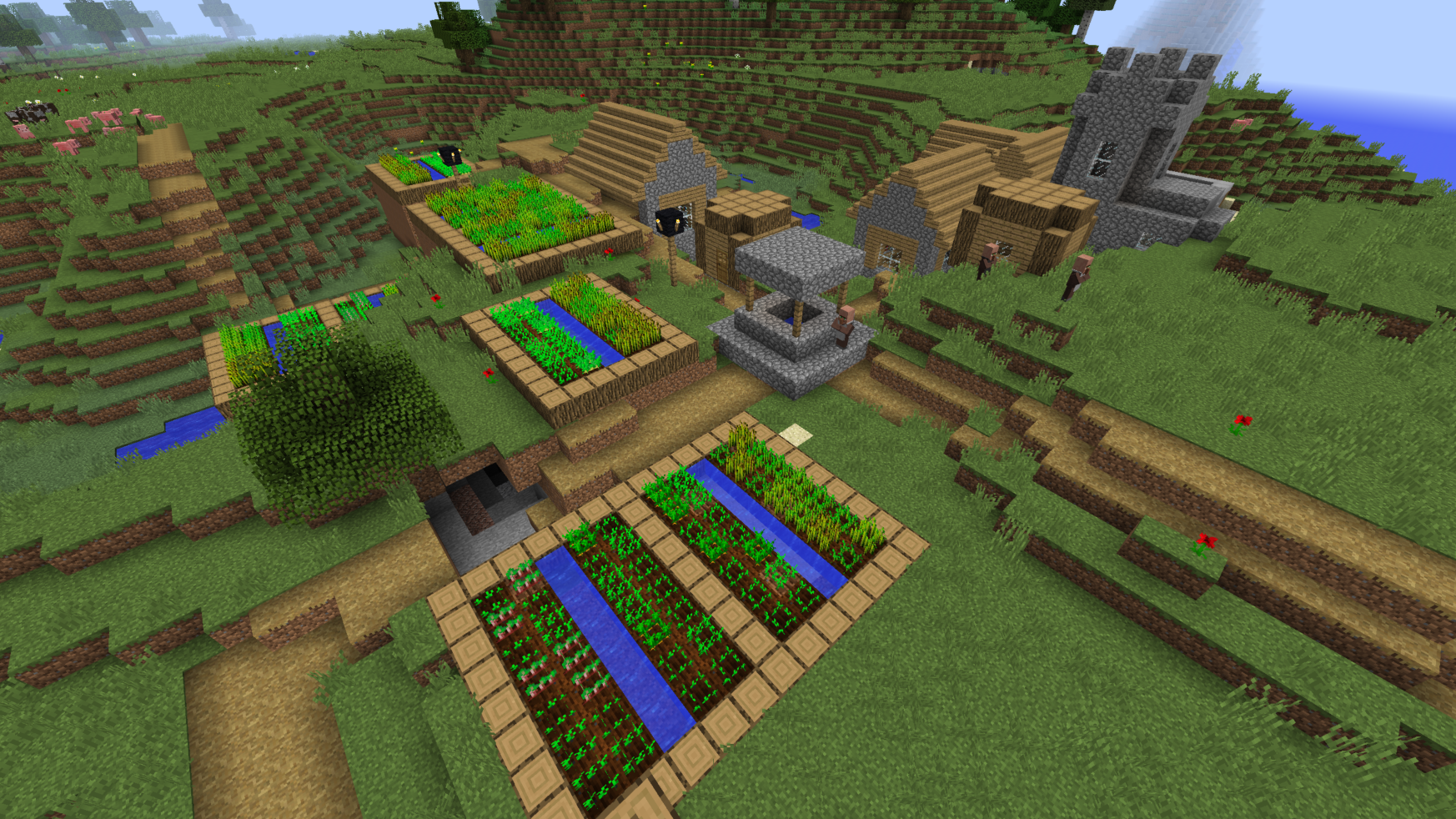}
  \caption{A typical Village generated by Minecraft's map generator. Because there is a limited selection of templates for the buildings you can already see repeated buildings. The road network is generated without taking the terrain into account, and is then projected on the terrain surface. This leads to inefficient and unbelievable layouts, with awkward elevation changes and roads being interrupted by large holes or chasms. }\label{fig:typicalVillage}
\end{figure}

\begin{figure}[tbh]
\includegraphics[width=\linewidth]{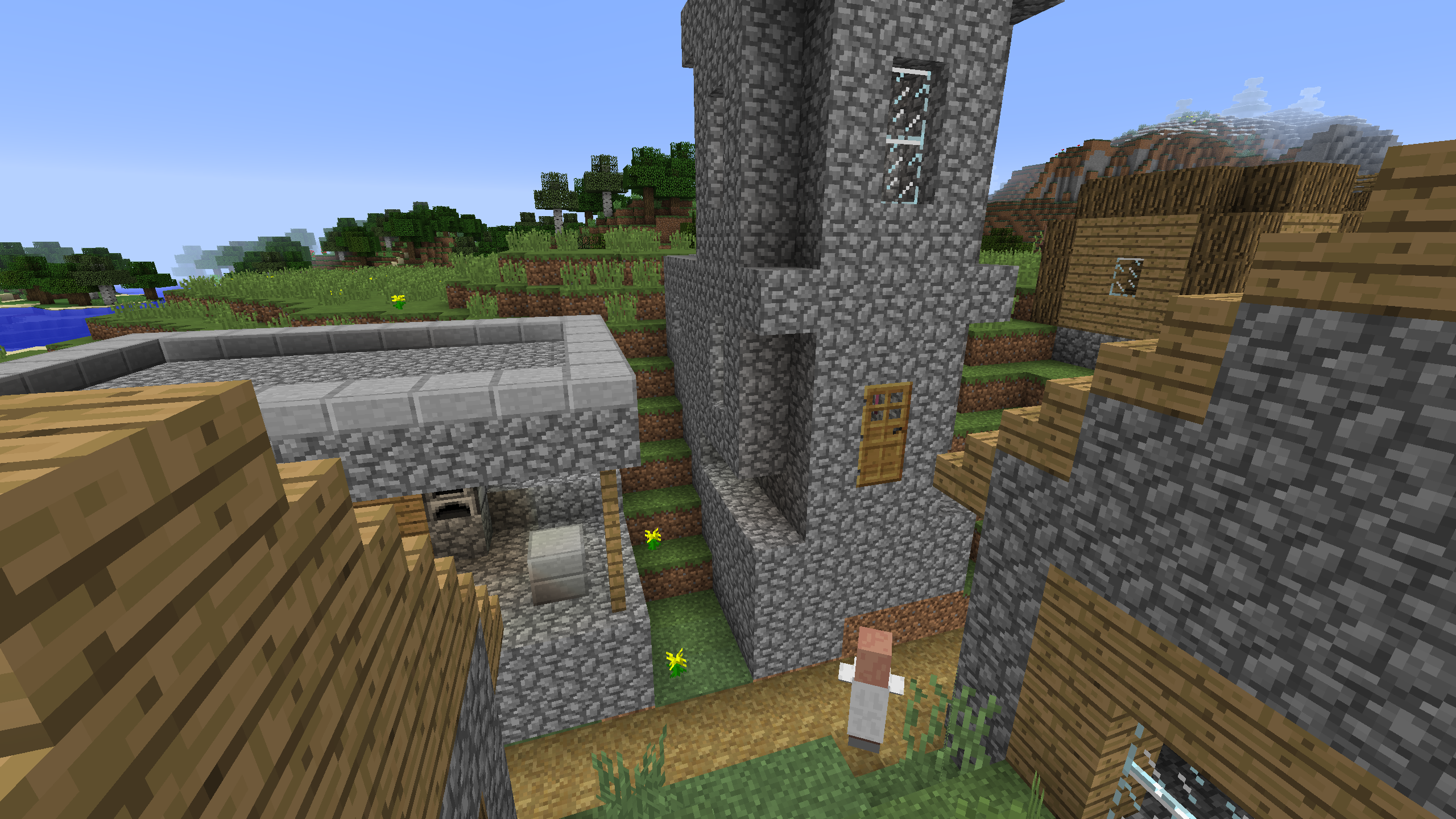}
  \caption{Building templates are placed on the average elevation of their supporting terrain. The empty blocks below buildings on a slope are filled in, but the building templates are not otherwise adapted. This leads to problems like the door of the church being several blocks above the ground level, which keeps the villager from entering it. }\label{fig:alignment}
\end{figure}

We believe Minecraft is a suitable testbed for this competition, for several reasons. Its popularity makes it accessible to many people, and there is a modding community with an interest in similar problems. Furthermore, Minecraft also comes with a built-in settlement generator (see Fig.\ref{fig:typicalVillage}), demonstrating that the problem is, at some level, solvable. However, if we compare the settlements generated by the Minecraft map generator with human made settlements in Fig.~\ref{fig:exampleSettlement}, we can see that there is a lot of room for improvement (see, for example, Fig.~\ref{fig:alignment}). We will discuss those missing elements in more detail in Sec.~\ref{Evaluation-Criteria}. Coincidentally, the elements missing are those that generative design and procedural content generation are interested in and challenged by \cite{colton2012computational,guckelsberger2017addressing,liapis2014computational}. So, submissions to the Minecraft settlement generation competition should be addressing some of the central challenges of PCG. 

\begin{figure*}[thb]
{
\begin{tikzpicture}[]
\node[anchor=south west,inner sep=0] at (0,0) {\includegraphics[width=\linewidth, height = 9cm]{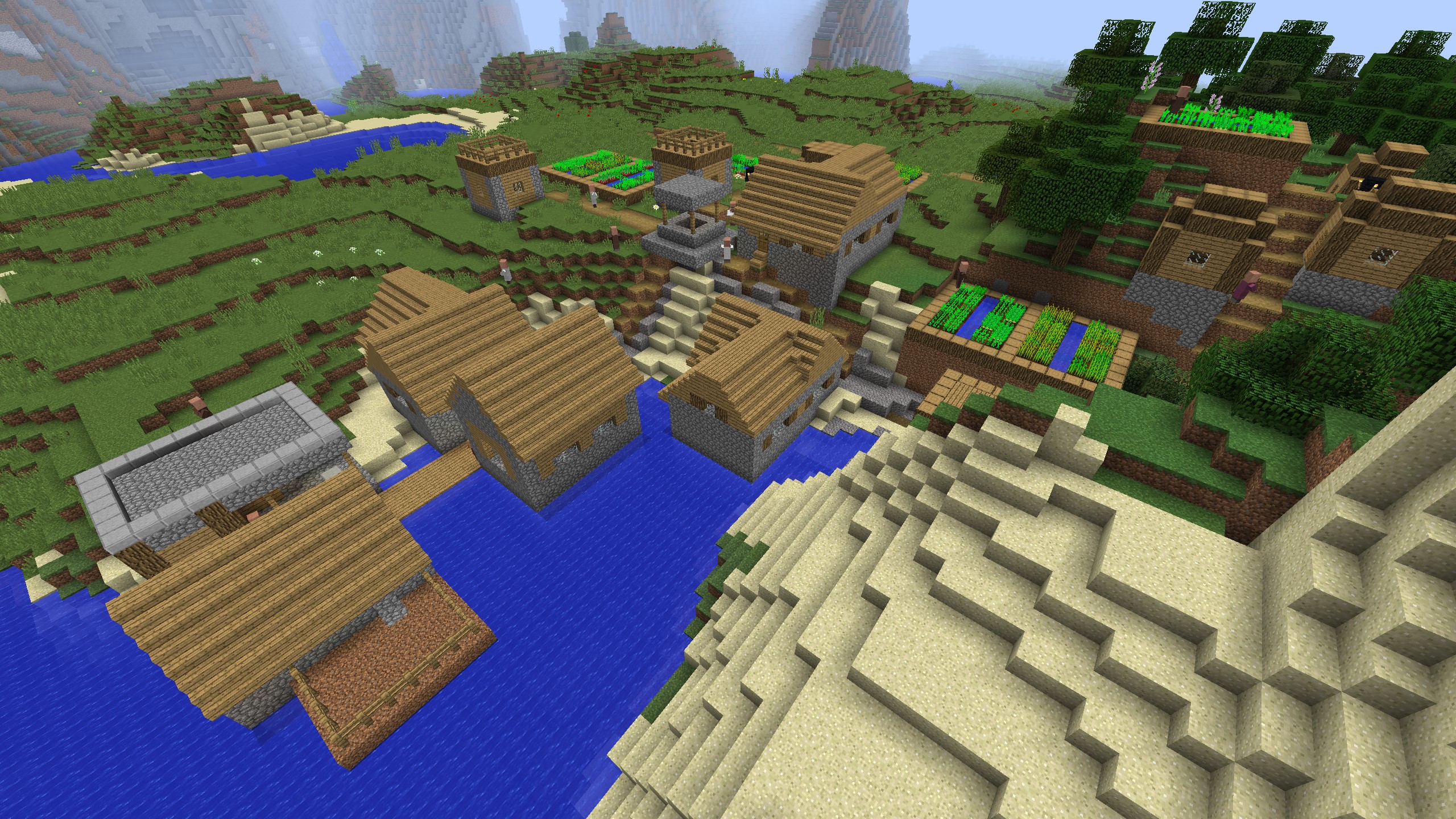}};
\end{tikzpicture}
}
\label{fig:problems}
\caption{Another typical village generated by Minecraft's map generator. Again, we see several problems. Several identical buildings, awkward elevation of the path leading to the middle well, and stone buildings placed directly in water. Also note the overall similar look to the village in Fig.~\ref{fig:typicalVillage}}
\end{figure*} 

\subsection{Aims}

This competition aims to encourage work in the area of procedural content generation that pushes for a.) greater \textit{adaptivity to content} and b.) towards \textit{holistic PCG} or orchestration. As both of these concepts are central to our motivation, we want to introduce them here:

\subsubsection{Adaptivity to Content.}
By adaptivity to content we refer to content generation that is responsive in regard to existing content (which can take many forms). Many PCG methods work in a ``vacuum'' - early approaches, such as L-systems~\cite{lindenmayer1968mathematical} or regular expressions, produce artifacts or content that is in no way responsive to outside influence. Randomness is used to introduce variation, and random seeds are sometimes used an an input, but the content produced by them is not responsive to the seed in any meaningful way, i.e. the seed is not guiding the algorithm towards specific content. 
A step up towards greater adaptivity is the use of input parameters. An example here is the usual approach to generating player specific content \cite{shaker2010towards}, where the player's behavior is analyzed and broken down into  a set of parameters, such as preferred difficulty, and the content is then created biased by those parameters. 
In this competition we want to push for even greater adaptivity, by encouraging algorithms that respond to rich, multidimensional input - in our specific case a three-dimensional map. While this input can be narrowed down to a set of parameters (for example, the biome type) there is a need to respond at least to the underlying topography of the height map. Furthermore, there is also the problem of implicit properties, such as defensibility or beauty, which are not immediately evident from the data. Extracting them is already a challenge and also introduces the further problem of an agent based perspective. A Roman general might look at a piece of terrain and think it makes for good defenses, while a modern city-planner might look at the same land and think of high building costs. So maps provide a rich, multidimensional input that also have implicit meaning, based on context and perspective. Being able to adapt to this kind of content is important to build settlements like a human would, but, more generally, this challenge is also important to address if we want to move towards social or co-creativity. During a possible follow up competition, it might be interesting to see if an algorithm can continue or complete a settlement started by a human. This would require an understanding of what the human meant or intended with their initial settlement. 

%\subsubsection{Holistic PCG}

\subsubsection{Holistic PCG}
Holistic PCG refers to the challenge of ensuring that the different types of generated content as well as the different dimensions in which content expresses itself fit with each other. While there are good algorithms to generate content for specific, well defined domains, such as trees, guns, dungeon layouts, music, etc, it is often problematic to combine the generate elements without supervision. A very similar concept has been expressed as the orchestration of creative facets \cite{antoniosliapis2015,liapis2014computational}, where facets can be visuals, music, and rules.

One approach is to ensure that the generated content is orthogonal, i.e. does not affect each other. For example, a game might generate the portrait and skill for an non-player character independently of each other, or a level and its background music might be generated separately. This still sometimes fails, and generates combinations that are unbelievable or inappropriate. This approach also often fails to produce interesting variations. Games sometimes advertise the fact that their procedural generation can create a million different artifacts of some type, meaning that they randomly generate 6 characteristics and each can have 10 states. In reality, this often feels more like 60 different kinds of content, because the combination of different characteristics do not interfere with each other (by design), and as a result, the overall ensemble has no emergent properties arising from a result of those combinations. 
More sophisticated approaches at orchestration employ hierarchical methods \cite{togelius2012compositional,smith2014logical}, using generators within generators, or passing parameters down a tree. For example, a game might first generate a landmass and then generate cities on that landmass that are influenced by the properties of the land. Those cities itself then have parameters, which in turn affect the generation of the buildings inside. Furthermore, in reality, the flow of causal influence is often circular, or as Liapis \cite{antoniosliapis2015} puts it, requires an ``iterative refining process''. Here again, settlements are a good example, as they often arise from a long history of interaction, in which different elements affect each other in turn, i.e. a city is shaped by its environment, and the cities environment is shaped by it in turn. This can be addressed by simulating the process that actually generates an artifact, rather than just trying to generate the artifact outright. This has the added benefit of grounding the artifact in an actual developmental history and giving it meaning, because it actually has a meaningful past. Unfortunately, this approach is usually computationally intensive, and poses the question of how fine grained the simulation should be. In any case, the task of holistic PCG, where all generated content facets are meaningfully related, is non-trivial. In the next section we will discuss the different aspects that have to be orchestrated in settlement generation, which should further illustrate why moving towards human-like settlement generation should also move us closer to holistic procedural content generation.

\section{Challenges for Human-level Settlement Generation}

In this section we discuss challenges specific to generating a Minecraft settlement on a given map. We think that those are some of the obstacles that need to be overcome to move closer to human-level design - both for this particular competition and for artificial creativity in general. We should point out, though, that this list is not exhaustive. It is likely that there are other challenges and obstacles to get to human-like settlement generation. Furthermore, we consider the challenges outlined here ambitious goals to move towards, and do not expect them to be fully solved right away. 

At the end of the chapter we also introduce the criteria that will be given to the human judges. They are derived from the challenges outlined in this section, because we want participants to address these problems. 

\subsection{Adaptation to Environment and Terrain}

Real-life human settlements are adapted towards their surrounding environment and terrain in multiple ways. On a per-building basis, houses and structures reflect the climate conditions they are built in. Furthermore, the environment of a settlement might also dictate the available resources for construction, and certain conditions might require specific types of buildings, such as irrigation, shelter, etc. A house in the cold climate of Edinburgh, for example, might be built sturdy, to protect against wind and cold and keep the warmth in. Likewise, the easy availability of stone might influence the choice of building material.   

Real-life settlements are also adapted to the immediate terrain they are in. Natural features, such as mountains or lakes, play a role in the selection of where to settle. Some terrain features can be both positive and negative for a specific settlement, and settlements are often built to take advantage of the benefits and to compensate for the problems. A river, for example, might provide more food and added mobility via ship travel, but might also limit mobility by foot, and be a potential flooding hazard. A settlement might subsequently be built close to the river (or even more likely, at a good location to cross the river), to take advantage of the benefits. The same settlement might also build a dam to protect against flooding, and houses close to the water might sit on stilts. If the settlement dumps waste into the river there might also be a development of different districts, where better, more desirable houses are further up on the river. 
 
On the other hand, humans change the terrain and environment in a limited fashion to better suit their needs. While it might be impossible to remove a mountain or an ocean, it might be possible to build a canal or tunnel. 

In summary, human settlements are shaped by and in turn shape the terrain and environment around them. If we study human-built settlements in Minecraft, we can see that these principles are realized in some existing builds. AI settlement generators, on the other hand, struggle to adapt to environment and terrain. They often rely on flattening large areas and then building settlements that do not reflect the surrounding map at all. Similarly, the houses and structures built by them are often templates that get placed as a whole unit, and are in no way reflective of the environment they are in. As part of this challenge, we encourage participants to develop generators that produce different settlements for different kinds of maps, reflecting the available materials, surrounding terrain and environmental conditions. In particular, successful entries should generate settlement that fit into the topography of the map without changing it too much, and build structures that fit into the topography. We also encourage participants of the challenge to think of other ways in which the given maps influences the generated settlement, ideally in such a fashion that it is evident to a human observer what kind of adaptation happened. Ideally, different maps should lead to different settlements, and it should be clear in which way the settlements where shaped by the map. Also note, that there is some overlap with the other three criteria, as their particular solutions are also often ideally adapted to the underlying map. 

\subsection{Functionality from an Embodied Perspective}

Minecraft is not just a creativity tool, but also a game. Subsequently, a Minecraft settlement is not just an aesthetic artifact; it also needs to fulfill certain functional roles. One big aspect of this is accessibility and mobility. Is it actually possible to walk through the settlement and reach its different parts? Are there doors into the buildings, are the bridges over the river? How easy is it to navigate the settlement? A second big function a settlement provides is protection from various dangers. Is the settlement well lit, or does it otherwise prohibit mobs from spawning? Does the settlement manage to keep dangerous mobs outside? Are there other forms of defenses and traps to protect the player? Finally, there is also a need to obtain food and process it - so is there an accessible way to produce food and not starve? This list of functional requirements is not exhaustive - it is just meant to give some examples.

This criterion does have a certain overlap with the challenge of adaptivity. Certain functionality only needs to be provided in some cases. For example, bridges compensate for reduced mobility introduced by obstacles. They are an adaptation to reduce the negative effects of the terrain. Other functionality takes advantage of given terrain features, such as river lock further enhancing the mobility of an existing waterway.

This should also be considered, not just from a player perspective, but also under the assumption that the NPC villagers living in the settlement have similar needs. There is crossover here with the next section, because some functionality displayed in typical human-build settlement is not functional in the strict sense.
For example, in reality a lot of houses have pitched roof to ensure better rain water runoff. In Minecraft, this is not necessary (as rain does not pool on the ground), but players still build houses with pitched roofs. This narrative functionality alludes to a functionality we understand, yet is not strictly functional in Minecraft. We still encourage the inclusion of such functional elements. Ideally, these narrative functionalities should also be adaptive to the environment. So, pitched roofs should be more popular in a rainy swamp than in a dry dessert, etc.    

In summary, the fact that there is an actual player in the world gives us an embodied perspective. The avatar can interact with the world in a variety of different ways, and a settlement can provide different affordances \cite{gibson1966senses,cardona2013cognitivist} to the player. The challenge in this competition is to produce a generator that can ensure that the settlement provides a maximum of functionality and affordances for the player, while also satisfying the other criteria. 
 
\subsection{Believable and Evocative Narrative}

Human settlements are not just collections of functional buildings, but they also tell a story about how they came about, who are the people living there, and how they see the world. When we look at human-created Minecraft settlements, we can often see how certain human or imagined cultures are reflected in the created buildings. There are cities that resemble ancient Rome, mythical Elven forest outposts, and modern US cities. Often, the buildings also reflect very clear narrative ideas, such as "this is a defensive mining outpost built in a harsh environment", or "this is a capital city, built to impress foreign and domestic visitors alike". These settlements are evocative, i.e. they manage to transport this story by looks alone. Relating to the earlier challenge of adaptivity, the better settlements also evoke narratives that work well with the terrain and biome they are in. The narrative they evoke should fit the terrain and environment of the given map, and ideally arise from it. Different maps should result in different narratives.

Human settlements also reflect how they came about. Medieval cores of modern cities, for example, tell a part of a settlement's origin story. This is something rarely seen, even in human-made Minecraft builds, as settlements here are created with the final state in mind. On the other hand, procedural content generation does have a tradition of simulationist approaches, which would be a possibility for this challenge. An AI could simulate people living and building a settlement over several stages, rebuilding or replacing structures, or slowly modernizing them. This could solve several of the previous problems, as buildings and the overall settlement would be the actual result of an adaptive process, and the produce of an actual sequence of events forming the basis for a possible narrative.

There is also a certain overlap between narrative and functional requirements. How easy the player can obtain wood for building and further process it is a functional requirement. To satisfy this a settlement could have a nearby forest and a workbench nearby. To make this work with a narrative of a settlement that relies on wood production it might also be good to build a structure that looks like a sawmill, and a street to transport the goods to town. This would not produce any additional functionality to the player (or the digital villagers), but it would tell a narrative that aligns with the functionality provided to the player. Similarly, there could also be elements that have purely narrative functionality, i.e. structures that have a believable fictional use that is not immediately reflected in the game mechanics. For example, a settlement could have an aqueduct to provide water, even though Minecraft characters do not need to drink.

To summarize, the particular challenge here is not just about narrative generation, but also about generating a narrative that fits with the generated settlement, and ensuring that this narrative is communicated to the player with the generated structures. Inspiration by existing, historical or imagined cultures can help to transport this narrative. The aim is to produce something that is both believable and evocative, while still being aligned with the previous criteria.  

\subsection{Visual Aesthetics}

Aesthetics are arguably subjective, yet architects and city planners usually follow a range of principles when it comes to designing a settlement or a house. While untrained humans manage to intuitively realize these principles with some success, it is difficult to design an algorithm with automated aesthetic judgment. Existing Minecraft mods that add structures to the world circumvent this problem by hand-designing appropriate templates, and then building a settlement out of those templates. This solution has several problems: it allows for little variation between the buildings, as they all need to be pre-designed, and it also allows for little adaptation of the buildings to the surrounding buildings, the underlying terrain, etc. Buildings are sometimes parameterized, and hence reflect some specific environmental settings, such as available materials or the climate they are in. This makes them more adaptive, but there seems to be trade-off between controlling the exact look of a structure, and making it adaptable towards uncontrolled environmental factors. The challenge in our case is to ensure that buildings still follow basic design principles while being adaptive, functional and evocative. 

There is also the further problem that the overall settlement itself should follow certain aesthetic rules. How buildings are aligned and what sight lines exist can play an important role for the overall feel of the settlement. Finally, there is also the question of how well the particular aesthetic expression chosen aligns with the other challenges. For example, a settlement which has a narrative of being a foreboding fortress, and is designed with that functionality, should also have an aesthetic that reflects this. 

\subsection{Evaluation Criteria} \label{Evaluation-Criteria}

Based on the different challenges we designed a set of criteria that will be used to evaluate the different algorithms. Each algorithm will be scored based on the settlements they generated for three different, unseen maps. The judges will award 0 to 10 points in each of four categories - Adaptability, Functionality, Narrative and Aesthetics. The judges will be provided with the following list of criteria to guide their evaluation for each category, with the understanding that this is a non-exhaustive list of what this criteria means. 

\subsubsection{Adaptability} 
\begin{itemize}
\item Do the structures in the settlement adapt to the terrain?
\item Do the structures in the environment reflect the environment, i.e. usage of available material, adaptation to the biome?
\item Does the settlement take advantage of terrain features or compensate for problems with the terrain?
\item Are the settlements different in reaction to the different initial maps?
\item Are there any other ways in which the settlement adapts to the given maps?
\end{itemize}

\subsubsection{Functionality} 
\begin{itemize}
\item Does the settlement provide protection from danger?
	\begin{itemize}
	\item Does it keep mobs from spawning?
	\item Does it keep mobs out?
	\item Protection from other environmental dangers?
	\end{itemize}
\item Is the settlement accessible to a player avatar in survival mode?
	\begin{itemize}
	\item Can you walk to everywhere?
	\item Does the settlement provide faster modes of transport?
	\item How easy is it to find your way around?
	\end{itemize}
\item Does the settlement provide the player with additional affordances?
\item Does the settlement make resources easy to obtain?
\item Is there an easy way to get food?
\item Does the settlement provide functionality to the villagers?
\item Does the settlement reflect the embodiment of the player avatar?
\item Is it appropriately scaled?
\end{itemize}

\subsubsection{Believable and Evocative Narrative}
\begin{itemize}
\item Is the settlement evoking an interesting story?
\item After looking at the settlement, could you give a short description of what this settlement is about that sets it apart from other settlements?
\item Is it clear what the function of the settlement is?
\item Does this function make sense in regards to the terrain and environment it is in? I.e. is the logging camp in a forest, the harbour town at the sea, ... ?
\item Is the functionality of the settlement supporting this narrative function? I.e. does the fortified frontier settlement have functioning walls, is the farming village equipped with functioning fields?
\item Does the final settlement give any indication of how the settlement developed? 
\item Is is possible to look at the settlement and imagine in what order things where built, or what stages the development of the settlement took? 
\item Is there an indication of the history of the settlement evident in the structure?
\item Are there any convincing and consistent allusions to human cultures or specific points in history that the settlement is modeled after
\begin{itemize}
\item Does the settlement have a culture - either fictional or historical - that is evident from the settlement?
\item Do you know things about this culture just by looking at the settlement?
\end{itemize}
\end{itemize}

\subsubsection{Visual Aesthetics}
\begin{itemize}
\item Does the settlement look good?
\item Is there a consistent look to the settlement? Does it appear that all structures belong to the same settlement?
\item Is there an appropriate level of variation in the existing structures?
\item Are there any jarring features that make the settlement look unbelievable?
\end{itemize}

\section{Competition details}

The following section outlines the details for the actual competition. Up to date rules and further information is available on our website\footnote{http://gendesignmc.engineering.nyu.edu/}. We will first introduce the framework we provide, and describe the existing example agent. We will then outline how submission will be evaluated.

\subsection{Framework} \label{framework}

Following the advise from Togelius \cite{togelius2016run} we wanted to keep the barrier-to-entry to participate low. We provide a framework (available for download on GitHub) \footnote{https://github.com/mcgreentn/GDMC} built on top of MCEdit \cite{mcedit}, an open-source map editor for Minecraft that enables the user to create and edit Minecraft map files. In MCEdit, you can create brand new maps from scratch, open saved maps, move around the map in the 3D viewer and modify it, and apply a \emph{filter} over a subsection of the map. A filter is a command associated with some effect, such as filling all the space with air (or any material), replacing one material for another, or building structures such as columns, walls, staircases or entire settlements, which is the ultimate goal of this competition.

A number of stock filters are included in the engine, and new filters can be made by writing a Python program describing the inputs or parameters of the filter (such as materials to be used, dimensions or orientations of the created structures, etc.) and the code to be executed, encapsulated in the \textit{perform} function.

To participate in the competition, competitors have to write a filter which generates a settlement. The filter file and additional needed libraries then form the basis for the submission. Participants can test the filter file themselves on a range of test maps, all within the same framework. We maintain a list of additional libraries online that will be included, and plan to extend it with reasonable request from participants. 

We evaluated different framework options, and chose this approach, as it a.) is easy to get started for participants b.) required relatively little development effort for us to in terms of interfacing with a Minecraft map and c.) allows us to streamline the evaluation of submissions. Our framework currently requires submissions to be written in Python, but we are in the midst of discussions to offer a submission option in Java. We imagine that there might be some further developments in terms of exact technical requirements, so we encourage participants to read our latest up-to date-rules on our website.

\subsection{Example Agent}

We included a simple example settlement generator with the framework called ``CASG'', short for ``Cellular Automata Settlement Generator''. Given a 3-dimensional space in the Minecraft world, it uses binary space partitioning~\cite{shaker2016procedural} to create \emph{yards}, building fence posts around the perimeter of each yard to designate the area in which buildings can be built. Within each yard, it selects a randomly sized rectangle, randomly finds four different heights for the four corners of the building, and builds columns to these heights. It then finds the average height between these columns and builds a ceiling at that level. The walls of the buildings are generated using cellular automata techniques (CA)~\cite{wolfram1983statistical}: first it randomly creates walls using a mix of glass and stone blocks, then several generations of CA are performed until they resemble windows.

\begin{figure}[tbh]
\includegraphics[width=\linewidth]{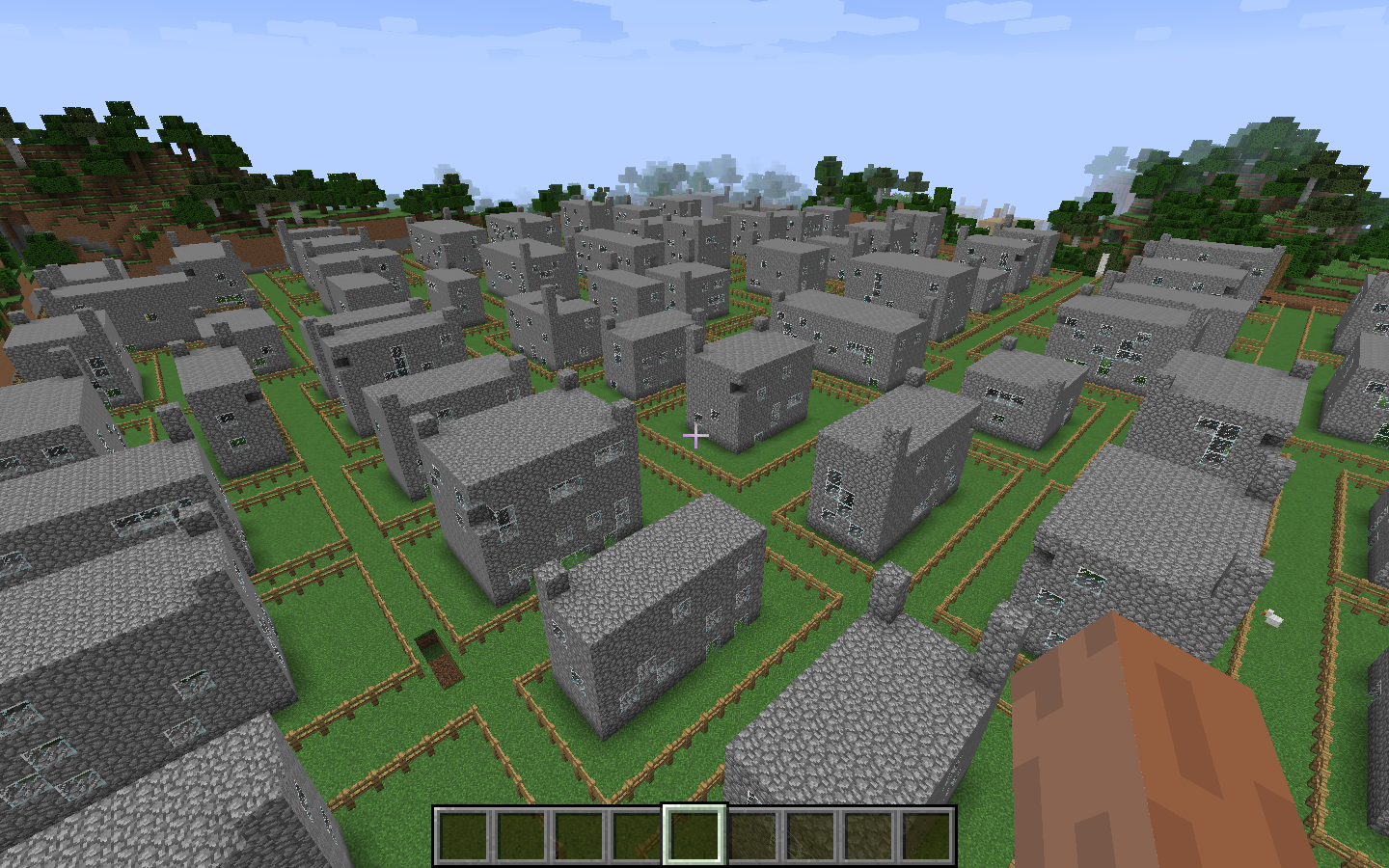}
  \caption{A settlement generated using the sample agent provided in the competition framework. This example uses Binary Space Partitioning to divide the terrain into lots, and a Cellular Automaton to generate the buildings.}\label{fig:Sample_Agent}
\end{figure}

\subsection{Submission}

Submission will be handled via the submission website and will consist of a file of code containing the algorithm (filter) that will create a settlement for each given input map. Authors will provide up to two pages of write-up detailing what techniques where used in the algorithm and what they perceive as its strengths and unique features. The write-up will not impact an agent's score, as judges will not have access to it during evaluation.

During submission the participants will be asked if they want to make their code publicly available. If they chose so, we will also publish their code on our competition website. In either case, we do not take ownership of the code, but we do encourage participants to make their code available to others. 

We do not, at this point, limit the submitted algorithms in terms of game based resources, such as how many blocks they can manipulate, or what resources are present on the map. We do require that algorithms do not run substantially longer than 10 minutes on a regular PC for a 256 by 256 map. This is to ensure what we can evaluate all maps in a reasonable time frame. 

\subsection{Evaluation}

The submitted algorithms will be judged based on a set of three different unseen maps. Example training maps will be provided beforehand. Every algorithm will be run on the same three unknown maps, and the resulting maps with settlements will be anonymized and given to the judges. The judges will also be able to see the 3 maps in their natural state, before the algorithm was applied.

The judges will then evaluate all 3 maps for a given algorithm by spawning at the central point (for the x and z coordinate) of the map. Although the judges will play the game in the Peaceful mode (where no monsters attack) to make exploration easier, they are encouraged to walk through the settlement from a Survival perspective. They are given a set of evaluation criteria (see Evaluation Criteria in Sec.~\ref{Evaluation-Criteria}), and are asked to assign points based on the settlements for the different maps. The sum of the points is the score of the algorithm. The algorithm with the highest score wins.

The plan is to have all judges evaluate all submissions. If this is not possible, due to high number of submissions, there might be several evaluation rounds, where earlier rounds are used to reduce the number of submissions by evaluating a smaller amount of maps with only a subsection of the judges.

The results will then be published online, along with the competition maps, the settlements each algorithm generated and their respective write-ups, foll all participants. If permitted, we will also make the code for each algorithm available online. 

\section{Related work}

This section introduces some techniques that have been applied to similar problems in the past. Some of the cited works might give participants a good starting point for their own work.

\subsection{Procedural Content Generation}

The field of procedural content generation has developed a range of techniques that might be useful for this task. General introductions are available \cite{katecompton2016,shaker2016procedural,short2017procedural}. Lopes and Bidarra~\cite{lopes2011adaptivity} also provide a survey on adaptivity in procedural content generation for games. 
While their focus is on adapting content to a player, the many application scenarios discussed therein provide a valuable source of inspiration.

%\subsection{A Taxonomy of PCG}

Shaker et al.~\cite{shaker2016procedural} propose a taxonomy of PCG, with seven dimensions that allow the comparison of solution, and we use it here to help frame the problem we are interested in:  generation for the competition happens \textit{offline} and \textit{automatically}, as the level is fully created before game play, with no human input after the start of execution of the algorithm and no modifications by the agent after the start of the game. Most content in Minecraft (including settlements) can arguably be classified as \textit{optional}, as the player is free to make their own goals and either ignore large portions of the map or build structures (such as bridges and ladders) to access otherwise inaccessible areas. This is in contrast to many games where level design would be considered necessary (for example, a single gap too wide in a platformer can make the whole game unplayable).

The starting landscape can perhaps be considered a \textit{dimension of control}, as varying the features of the terrain, such as its biome and flatness or roughness of the terrain should impact the result of the algorithm. Other dimensions of control (such as random seed or parameters) can be explored by the designer, but it should be noted that the filters are expected to run during evaluation with whatever default parameters are implemented. Content is \textit{adaptive}, but unlike most examples of adaptive content that adapt to a particular player and their behavior, our settlements are expected to adapt to the provided starting map. The idea of environment-adaptive PCG highlights the limits of this taxonomy, as the input is much richer than a parameterization of a typical content generator; it is more akin to the input data in a data game~\cite{friberger2013data}.

While we welcome either \textit{stochastic} or \textit{deterministic} methods, a single settlement will be generated from each map for each agents, so that all judges will evaluate the same group of settlements for each applicant. The choice of \textit{constructive versus generate-and-test} methods is also left to the designers. Some considerations of how direct or simulation-based evaluation functions could be used for a generate-and-test approach are explored in section~\ref{generate-and-test}.

It is interesting to note that Minecraft in general, and settlement generation in particular are not perfect fits for classification in some of these dimensions (such as necessary vs optional and adaptivity), due to the unique characteristics of the problem at hand. 

\subsection{Evaluation Methods for PCG Methods} ~\label{generate-and-test}

Another approach to this problem would be to write a generator with a larger expressive range \cite{smith2010analyzing}, and then test if the generated settlement fits certain requirements. When it comes to evaluating generated content, the search-based PCG framework differentiates between three kinds of evaluation functions~\cite{yannakakis2011experience}:
\begin{itemize}
\item \textbf{Direct evaluation functions} 
\item \textbf{Simulation-based evaluation functions}
\item \textbf{Interactive evaluation functions}
\end{itemize}
In interactive evaluation, content is evaluated during game play by a human user. 
This is the form of evaluation used to judge the competition, due to the openness of the problem. Obviously, this approach cannot be used during the actual, unsupervised settlement generation. Nevertheless, it would in theory be possible to generate a lot of settlements before submission and evaluate them with human input. This could then form the basis of a data set used for training an algorithm such as a neural network. The trained network could then be submitted as a filter. 

Alternatively, the algorithm could use direct evaluation based on some  desirable design characteristics. These could be captured by metrics such as number of building or roads or mean distance between buildings, and could then be directly evaluated during the execution of the algorithm.

A simulation based evaluation is also an option. For example, an AI-controlled character could be designed (e.g. by using the Malmo framework \cite{johnson2016malmo}) to explore the environment and test if certain areas are accessible, certain resources are reachable, etc. Tanagra~\cite{smith2010tanagra}, Ropossum~\cite{shaker2013ropossum}, Sentient Sketchbook~\cite{liapis2013sentient} are examples of game authoring tools using simulation or direct methods of evaluation of content.

Competitors could use these approaches to see if their algorithms generate sensible content on a large number of random maps, increasing the confidence that it would perform well in unseen maps. Clearly, designing such metrics or agents is a considerable challenge in itself, so it will be interesting to see if competitors will employ any algorithm techniques of evaluation,  and how the development of more complex generators affect this decision in future installments of the competition.

\subsection{City generation}
The field of computer graphics has a range of existing approaches to automated city generation, such as Kelly and McCabe's \textit{Citygen}~\cite{kelly2007citygen,kelly2008interactive}. They combine several existing techniques to generate different elements. Similar to~\cite{muller2006procedural} they use L-systems~\cite{lindenmayer1968mathematical} to generate roads and buildings. Perlin Noise~\cite{perlin1985image} is used to generate terrain.
Parish et.al.~\cite{parish2001procedural} provide another example of a city generator, which integrated real world data to obtain street and building placement. Groenewegen et.al.~\citep{egs.20091045} introduce an alternative approach that generates city subdivisions based on land usage model and the underlying terrain. Both approaches realize a form of adaptation to existing content and allow the user to influence the generated city. 

There are also several Minecraft mods, player generated modifications for the game, that touch on settlement generation. Millenaire \cite{millenaire} is a mod focused on settlement diversity. It adds a range of villages for different cultures and with different functions, such as farming village or fortress. It also massively enhances the functionality of settlements: villagers collect resources to build new buildings to expand their settlements. Buildings are constructed block by block, by a villager avatar. This system is interactive to an extent, as the player can sell resources to villagers to support their buildings projects. The constructed settlements are also somewhat adaptive. For example, the pathways in the settlement are created by connecting different buildings along the shortest path, determined with an A* algorithm. Another mod, named Lost Cities \cite{lostcities}, changes Minecraft's world generation so the whole world is covered in city ruins. 

There are also other games that feature automatic settlement generation. Ultima Ratio Regum \cite{URR} is a game that heavily relies on procedural content generation to create its gameworld. It is a prime example of hierarchical generation, as it first generates nations and their cultures, and then generates cities, buildings, religions and speech to fit into these cultures.   

\subsection{Simulation as Generation}

Rather than using simulation to evaluate generated content it is also possible to simulate the processes that generate content in the first place. For example, \emph{Dwarf Fortress} \cite{dwarffortress} generates hundreds of year of backstory, including biographies of historical figures, by basically simulating hundreds of years of history (to a certain level of abstraction). This has the advantage that the generated artifact is usually consistent and more meaningful, as it is the product of an actual process of creation. A possible refinement here is using a agent based simulation, where the process is simulated by an actor with an agent centric perspective and agent specific motivations. Imagine a bunch of villagers, with their respective needs, building houses and changing the world in order to survive. While these ideas have been discussed for games, as far as we know, there is no actual game that uses this approach.

The advantages of creating content based on actual agent adaptation are discussed by Guckelsbeger et.al.~\cite{guckelsberger2017addressing}. There are also some research prototypes which explore the idea of using agents driven by intrinsic motivation to change their environment \cite{salge2014changing}. Their experiments are done in a simplified Minecraft-like world, and they argue that this approach allows for artifacts that reflect the embodiment of the agent. They also raise the issue of scalability, which seems to be a large obstacle for their, or any other approach for simulation as generation.

\section{The Road Ahead}

We named the overall competition framework ``Generative Design in Minecraft'' because we hope that the Settlement Generation Competition will be a first step towards more challenging competitions that push the ideas of adaptive and holistic PCG even further. And while a lot depends on the level of participation and what problems prove to be hard and easy, we developed a tentative road map on where we might see these competitions go in the future. 

One path would lead us further towards co-creativity and collaborative generation. In the current competition, the algorithms only have to adapt to a given terrain map, but this could easily be extended by adding existing human-made settlements to the map. In this case the algorithms task would be to expand and complete the settlements, ideally in a way that complements the existing settlement. This would touch upon a range of interesting questions, starting with style transfer and going all the way towards understanding the intentions of the original builders of the existing settlement from the existing content.

This could be made even more challenging, by changing the framework to something like project Malmo, and situating the competition algorithms inside the game. Their task would then be to interactively assist a human who is building a settlement in real time. 

Another path ahead would be to push to even greater integration between different kinds of content. Minecraft does offer the possibility to have books with actual text in the world. One way to use this would be to ask an algorithm to not only create an evocative settlement that tells an interesting story, but to produce that story as a written narrative as well - a town chronicle that is stored somewhere in the town itself. Another option to integrate written word with world generation would be the task of automated treasure hunt generation. Could an algorithm create a world that hides a treasure and clues to this treasure, and then write a book that leads a player on a merry treasure chase?

%\section{Conclusion}

\begin{acks}
CS is funded by the EU Horizon 2020 programme under the Marie Sklodowska-Curie grant 705643. RC gratefully acknowledges the financial support from Honda Research Institute Europe (HRI-EU). MCG would like to thank the GAANN program for his funding.
\end{acks}

\bibliographystyle{ACM-Reference-Format}
\bibliography{sample-bibliography}

\end{document}